\documentclass[runningheads]{llncs}
\usepackage[T1]{fontenc}
\usepackage{hyperref}
\usepackage{graphicx}
\usepackage{booktabs}
\usepackage{multirow}
\usepackage{colortbl}
\usepackage{esvect}
\usepackage{ifthen}
\usepackage{appendix}
\usepackage{graphicx}  % 用于插入图片
\usepackage{float}  % 控制图片浮动位置

\usepackage[table]{xcolor}

\usepackage[most]{tcolorbox}

\tcbset{
  enhanced,
  boxrule=1.5pt,
  colframe=gray,
  colback=white,
  coltitle=white,
  fonttitle=\bfseries,
  arc=3mm,
  boxsep=1mm,
  toptitle=1mm,
  bottomtitle=1mm,
  colbacktitle=gray,
  width=12.3cm,
}

\definecolor{darkred}{RGB}{200, 0, 0}
\definecolor{darkyellow}{RGB}{250, 153, 0}
\definecolor{darkgreen}{RGB}{0, 180, 0}

\definecolor{green1}{RGB}{230, 245, 230} % 浅绿色
\definecolor{green2}{RGB}{200, 230, 200} % 中等绿色
\definecolor{green3}{RGB}{150, 210, 150} % 深绿色
\definecolor{red1}{RGB}{245, 230, 230} % 浅红色
\definecolor{red2}{RGB}{230, 200, 200} % 中等红色
\definecolor{red3}{RGB}{210, 150, 150} % 深红色

\begin{document}
\title{Rethinking the Chain-of-Thought: The Roles of In-Context Learning and Pretrained Priors}
%
% \titlerunning{Rethinking the Chain-of-Thought: The Roles of In-Context Learning and Pretrained Priors}
\author{Hao Yang\inst{1} \and
Zhiyu Yang\inst{2} \and Yunjie Zhang{3} \and Shanyi Zhu{4} \and
Lin Yang\inst{1}\thanks{Corresponding author}\orcidID{0000-0001-9056-0500}}
\institute{School of Intelligence Science and Technology, National Key Laboratory for Novel Software Technology, Nanjing University \\
\email{howyoung80@163.com, linyang@nju.edu.cn} 
\and School of Computing and Information Systems, Singapore Management University 
\email{kelvin.yangzhiyu@outlook.com} 
\and Central South University 
\and School of Global Education and Development, International Chinese Language Education, University of Chinese Academy of Social Sciences\\ \email{2413589021@qq.com}}

\maketitle              % typeset the header of the contribution
\begin{abstract}
Chain-of-Thought reasoning has emerged as a pivotal methodology for enhancing model inference capabilities. Despite growing interest in Chain-of-Thought reasoning, its underlying mechanisms remain unclear. This paper explores the working mechanisms of Chain-of-Thought reasoning from the perspective of the dual relationship between in-context learning and pretrained priors. We first conduct a fine-grained lexical-level analysis of rationales to examine the model's reasoning behavior.
Then, by incrementally introducing noisy exemplars, we examine how the model balances pretrained priors against erroneous in-context information. Finally, we investigate whether prompt engineering can induce slow thinking in large language models. Our extensive experiments reveal three key findings: (1) The model not only quickly learns the reasoning structure at the lexical level but also grasps deeper logical reasoning patterns, yet it heavily relies on pretrained priors. (2) Providing sufficient exemplars shifts the model’s decision-making from pretrained priors to in-context signals, while misleading prompts introduce instability. (3) Long Chain-of-Thought prompting can induce the model to generate longer reasoning chains, thereby improving its performance on downstream tasks.
% \footnote{Our code  and data will be publicly accessible when the paper is accepted.}
\end{abstract}
% \vspace{-12pt}
\section{Introduction}
\vspace{-8pt}
As models scale up, large language models (LLMs) exhibit emergent \textbf{I}n-\textbf{C}ontext \textbf{L}earning (ICL) capabilities~\cite{fewshotgpt}. ICL enables LLMs to perform tasks by leveraging a few exemplars presented as demonstrations. Relying on ICL, \textbf{C}hain-\textbf{o}f-\textbf{T}hought (CoT) prompting incorporates rationales into exemplars, guiding models to solve problems step by step and significantly improving performance across various downstream tasks~\cite{cot}. The reasoning process in CoT prompting can be formalized as:
\vspace{-6pt}
\[
\text{Input: } X = \{(q_i, r_i, a_i)\}_{i=1}^{N},\; q
\]

\begin{figure}[h]
\centering
    \includegraphics[width=0.85\linewidth, height=0.35\linewidth]{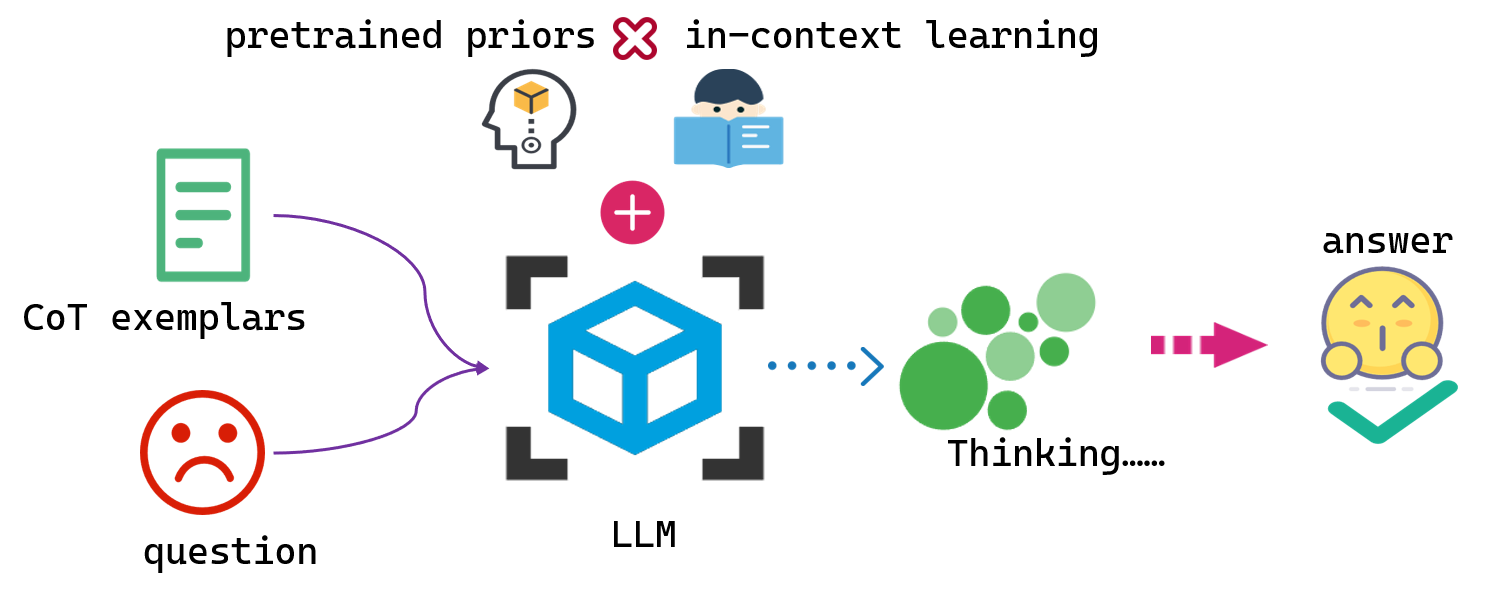}
    \caption{Framework highlighting the synergy between \textcolor{darkyellow}{\textbf{Pretrained Priors}} and \textbf{\textcolor{darkgreen}{In-Context Learning}} in powering CoT reasoning.
    }
    \label{fig:intro}
\end{figure}
where \( (q_i, r_i, a_i) \) represents the \( i \)-th exemplar consisting of a question \( q_i \), a rationale \( r_i \), and an answer \( a_i \); \( N \) denotes the number of exemplars; and \( q \) is the target question. The model generates a response \( (r, a) \) by conditioning on the exemplars and the question:
\[
P(r, a \mid X, q) = P(r \mid X, q) \cdot P(a \mid X, q, r)
\]

CoT prompting has become an essential technique for enhancing model reasoning capabilities. Previous studies have extensively explored how CoT affects reasoning performance, focusing on factors such as difficulty, step length, the number of exemplars, and order~\cite{contrastive,steplength,cot,cotlength2025}. 

However, the underlying mechanism of CoT prompting remains inconclusive. Inspired by~\cite{lin2024dual,pan-etal-2023-context}, this paper explores the dual relationship between ICL and pretrained priors in CoT reasoning. In studies on the mechanism of ICL,~\cite{min-etal-2022-rethinking} maintained that models primarily relied on pretrained priors during inference without acquiring new task-specific knowledge. This view was reinforced by~\cite{si-etal-2023-measuring,kossen2024context}, who argued that LLMs struggle to overcome their pretrained preferences. In contrast,~\cite{pan-etal-2023-context,wei2023larger} demonstrated that model scaling enables the learning of novel input-label mappings, although this capability emerged only in larger models (exceeding 66B parameters) rather than in smaller ones. Regarding CoT mechanisms, current research presents divergent perspectives: \cite{invalid} proposed that covariates in the prompts other than logical reasoning may be responsible for the performance improvements, while~\cite{counterfact} attributed its effectiveness to task simplification achieved by decomposing complex problems. Furthermore, ~\cite{madaan2022textpatternseffectivechain} suggested that models primarily imitated the CoT format through pattern recognition from exemplars, and ~\cite{merrill2024the} provided theoretical analyses of how CoT amplifies models' computational capabilities. Although these hypotheses require further empirical validation, they collectively advance our understanding of CoT's fundamental mechanisms and offer valuable directions for future research.

% As shown in Fig.\ref{fig:intro}, we investigate the interaction between ICL and pretrained priors in CoT reasoning. Our study firstly addresses two primary questions: \textbf{(1) What do LLMs learn from CoT exemplars through ICL? (2) Can LLMs override pretrained priors through ICL?} Based on our experimental findings, we further explore \textbf{(3) whether prompt engineering can leverage ICL and pretrained knowledge to elicit slow thinking}.

As shown in Fig.\ref{fig:intro}, we investigate the interaction between ICL and pretrained priors in CoT reasoning. Our study addresses three key questions: \textbf{(1) What do LLMs learn from CoT exemplars through ICL? (2) Can LLMs override pretrained priors through ICL? (3) Can prompt engineering leverage both ICL and pretrained knowledge to elicit slow thinking?} By disentangling the roles of ICL and pretrained priors, our findings shed light on the mechanisms behind reasoning emergence in LLMs, offering both theoretical insights into model cognition and practical implications for designing more effective prompts.

To answer these questions, we designed a series of experiments analyzing how LLMs integrate CoT exemplars and pretrained priors in decision-making. First, we provided models with both task-specific and task-agnostic exemplars to examine how variations in input influenced reasoning behavior. Through a fine-grained input-output analysis at the lexicon level, we quantified the reliance of LLMs on ICL signals versus pretrained priors. To assess robustness, we introduced controlled noise into exemplars and progressively increased the number of noisy samples to observe model behavior under misleading information.  These previous experimental results lead to the third research question: given both ICL signals and pretrained priors, can prompt engineering guide models to generate longer CoT rationales? To test this, we applied prompts that encouraged slow thinking, aiming to generate extended CoT rationales and improve downstream performance. Our findings revealed three key insights:
\vspace{-4pt}
\begin{itemize}
    \item LLMs learn reasoning patterns from in-context exemplars, while CoT reasoning still remains influenced by pretrained semantic priors.
    \item Smaller models (8B) can map inputs to outputs via CoT prompting, with an increased number of exemplars shifting reliance from pretrained priors to ICL signals. Additionally, low-quality exemplars increase instability.
    \item Prompt engineering can induce longer CoT outputs, improving downstream performance and suggesting a path toward model self-evolution.
\end{itemize}

\vspace{-10pt}
\section{Experimental Setup}
\subsection{Evaluation Tasks} 
Given CoT's pronounced impact on reasoning tasks, we focus on arithmetic, commonsense, and symbolic reasoning, following~\cite{contrastive,steplength,counterfact,cot}. For arithmetic reasoning, we employed GSM8K~\cite{gsm8k} and MATH-500~\cite{math-500}, two widely adopted benchmarks in mathematical reasoning. For commonsense reasoning, we used Date Understanding~\cite{bigbench}, which features compositional questions designed to assess commonsense. For the symbolic reasoning task, we considered two tasks: the Coin Flip task~\cite{cot} and the Last Letters Concatenation (four words) dataset~\cite{zeroshotcot}. The Coin Flip task assesses the ability to reason about probabilistic outcomes, while the Last Letters Concatenation dataset evaluates the capacity to manipulate and reason about symbolic representations.
\vspace{-6pt}
\subsection{Models \& Prompts}
\paragraph{\textbf{Models.}}
We conducted experiments using publicly available pretrained models of various series and sizes, including Gemma2-9B, Gemma2-27B~\cite{gemmateam2024gemma2improvingopen}, LLaMA3.1-8B~\cite{grattafiori2024llama3herdmodels}, and Qwen2.5-32B~\cite{qwen2.5}. Following~\cite{cot}, we applied greedy decoding. Due to space constraints, only a subset of the results is presented in the main text.
\vspace{-6pt}
\paragraph{\textbf{Prompts.}}
We employed standard and few-shot CoT exemplars from~\cite{cot} and a zero-shot CoT instruction (``\textit{Let's think step by step.}'') from~\cite{zeroshotcot}. Each dataset is associated with a task-specific CoT prompt. When a CoT prompt designed for one task is applied to a different task, it is referred to as a task-agnostic prompt. For example, in Section~\ref{subsection:rq1}, we utilized task-agnostic prompts (e.g., Sports-CoT, Date-CoT, etc.) on the GSM8K dataset.
\vspace{-6pt}
% \paragraph{\textbf{Evaluation Metrics.}}
% We extracted answers from model outputs using regular expressions and evaluated reasoning capabilities with accuracy metrics by comparing the generated answers against the ground truth.

\vspace{-6pt}
\section{Experiment}
% \vspace{-2pt}
\subsection{RQ1: What do LLMs Learn from CoT Exemplars Through ICL?}
\label{subsection:rq1}
This study compares CoT reasoning acquired through pretraining with that learned from exemplars through ICL, revealing what models extract from exemplars and the characteristics of pretrained CoT reasoning. 
% To avoid the influence of post-training fine-tuning data,
We evaluated base models of various scales and series, including Gemma2-9B, Gemma2-27B, and LLaMA3.1-8B.
\vspace{-10pt}
\begin{figure}[h]
\centering
   \includegraphics[width=0.7\linewidth, height=0.3\linewidth]{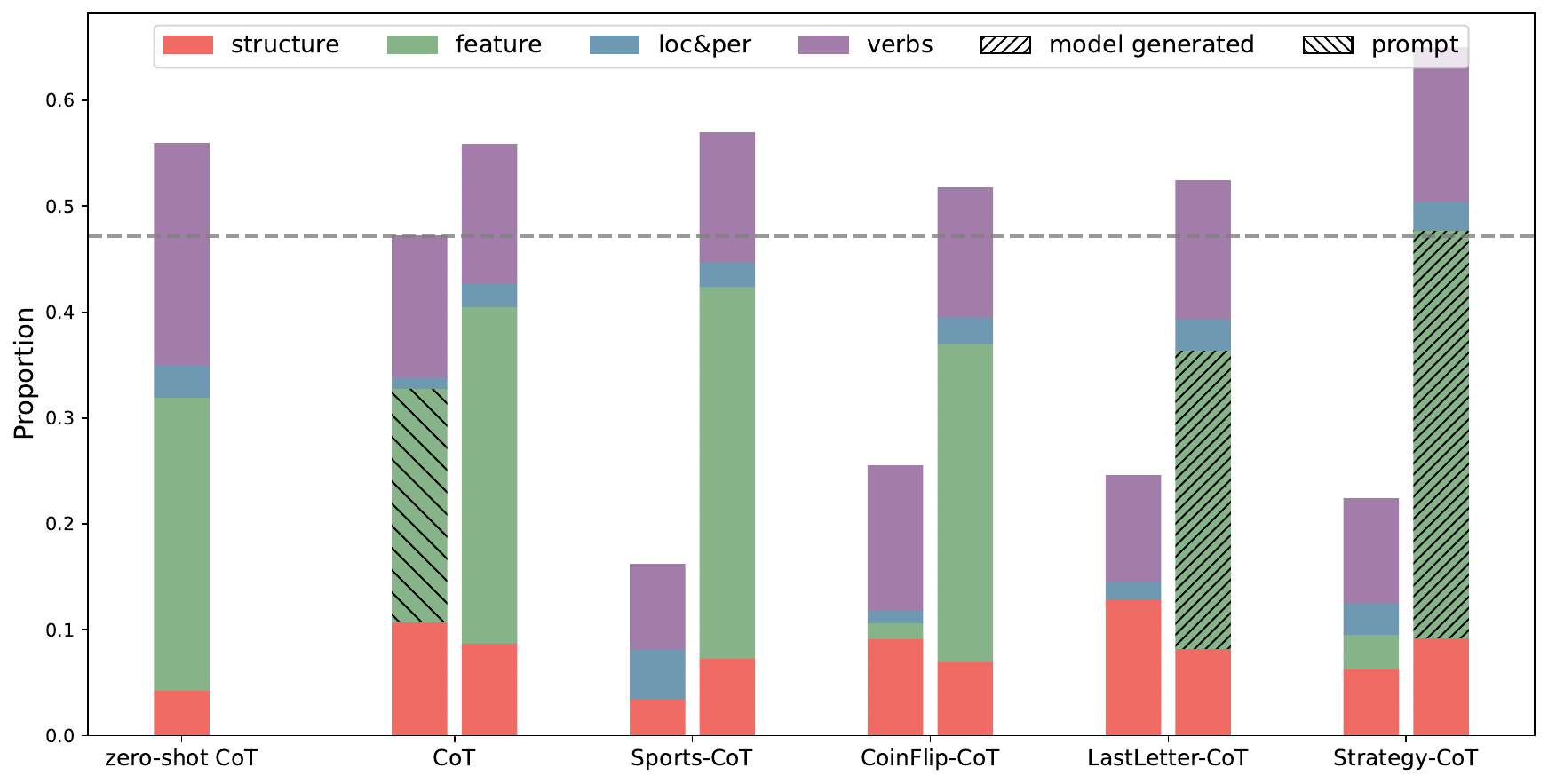}
   \vspace{-5pt}
   \caption{Proportion analysis of rationale components: 
\textcolor{darkyellow}{\textbf{Exemplars}} (left) compare in-context \textbf{CoT} vs. \textbf{task-agnostic CoT} variants (Sports, Coin Flip, etc.); 
\textcolor{darkgreen}{\textbf{Model-generated}} content (right) contrasts \textbf{zero-shot CoT}, \textbf{CoT}, and \textbf{task-agnostic CoT} frameworks.}
    \label{fig:word}
\end{figure}
\vspace{-26pt}
\paragraph{\textbf{Method.}} From an ICL perspective, we conducted a fine-grained analysis of model-generated reasoning texts using the GSM8K test dataset. We decomposed exemplars into four key components: structure words, feature words, verbs, and location \& person (loc\&per) entities. Structure words capture textual flow (e.g., ``The'', ``So'', ``Therefore'', ``Then'', ``Thus''). Feature words highlight task-specific elements such as numbers and operators (``+'', ``-'', ``*'', ``/'') in mathematical reasoning, extracted via string matching. Verbs represent reasoning actions, which we identified using the NLTK toolkit\footnote{\url{https://www.nltk.org/}}. Loc\&per words, also extracted with NLTK, denote entities related to location and individuals. We then explored model inference using varied prompts: zero-shot CoT, CoT, and  task-agnostic CoT. The zero-shot CoT prompt reveals the model's pretrained preference for CoT reasoning, while CoT and task-agnostic CoT prompts reflect the influence of in-context exemplars. This analysis provided insights into how different prompts shape reasoning behavior.
% \vspace{-20pt}

% \vspace{-10pt}
\paragraph{\textbf{Analysis.}} Fig. \ref{fig:word} highlights distinct differences in inference content across the three prompt types. First, even with task-agnostic CoT, the model continued to perform mathematical reasoning based on the specific question. The notable rise in mathematical feature words indicates that task-agnostic  CoT prompts did not alter the underlying reasoning behavior; LLMs remained guided by pretrained priors. Second, the CoT and task-agnostic CoT exemplars led to a significant increase in structural vocabulary compared to zero-shot CoT, suggesting that the model readily adopted lexical structures and mimicked exemplar language patterns. Finally, zero-shot CoT yielded a higher verb count than CoT. Under CoT and task-agnostic CoT prompts, verb usage declined. Given that verb frequency reflects sentence dynamism and causal expression~\cite{verb3,verb2,verb1}, this finding suggests a deeper-level imitation of exemplar reasoning forms. We further analyzed verb usage to explore this effect.
% \vspace{-25pt}
\begin{figure}[h]
\centering
\includegraphics[width=0.28\linewidth]{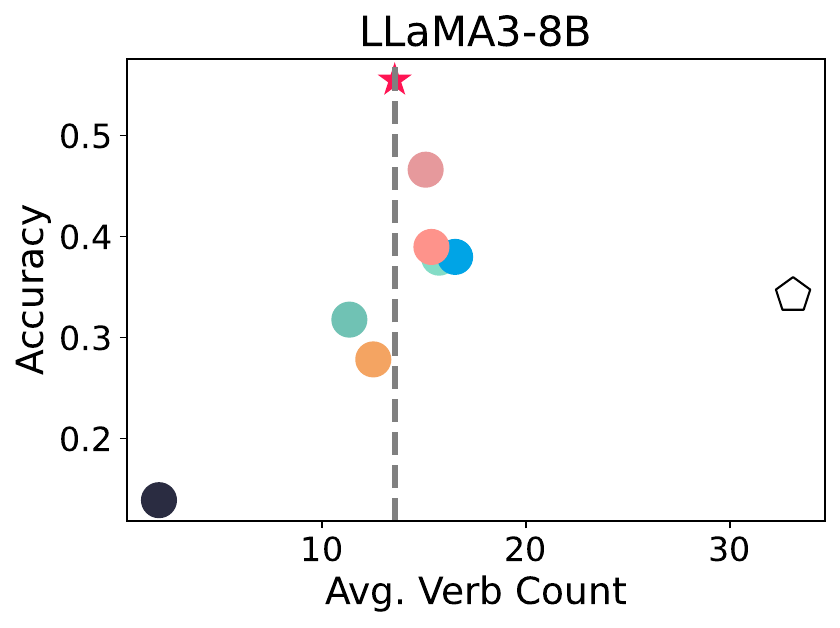}
    \includegraphics[width=0.28\linewidth]{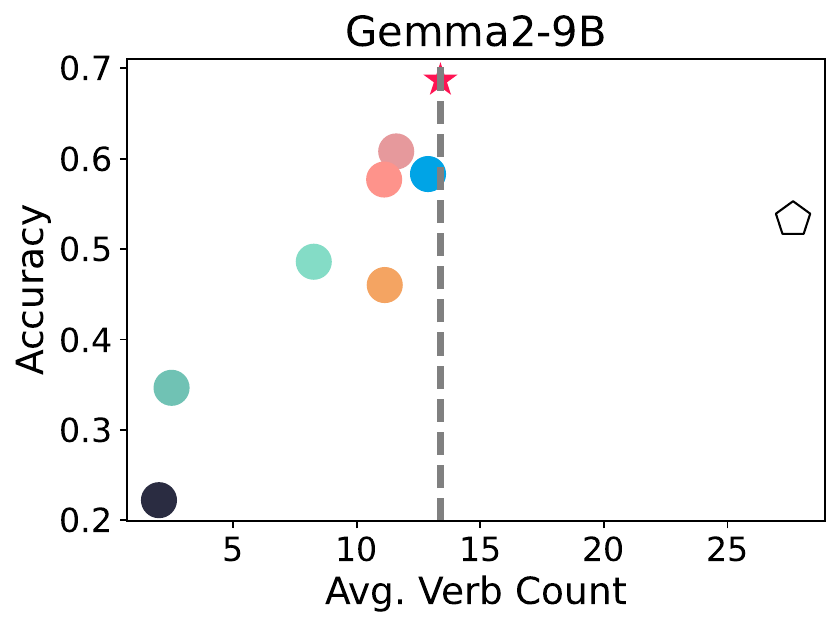}
    \includegraphics[width=0.28\linewidth]{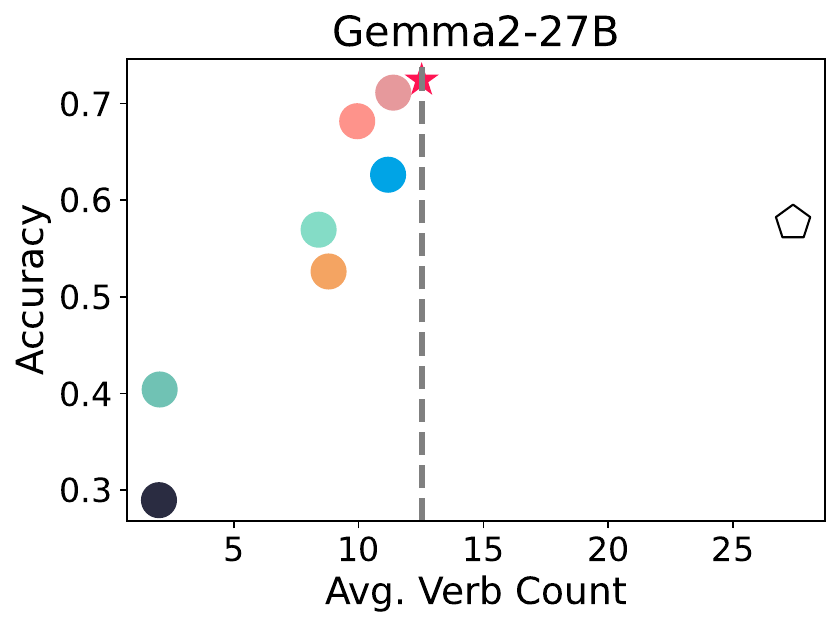}
    \includegraphics[width=0.1\linewidth]{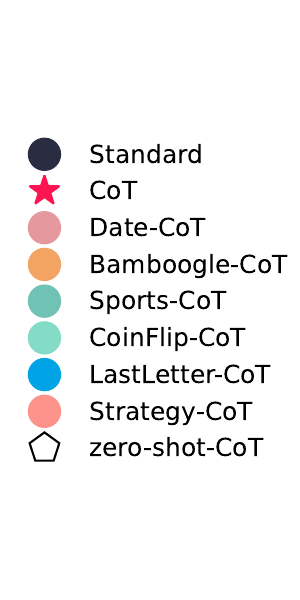}
    \caption{Scatter plot illustrating the relationship between the average number of \textcolor{darkgreen}{\textbf{reasoning verbs}} in model-generated content and \textcolor{darkyellow}{\textbf{accuracy}} under \textbf{CoT} and \textbf{task-agnostic CoT} settings.}
    \label{fig:action}
\end{figure}
% \vspace{-15pt}
Building on~\cite{steplength,lee2025llmscompresschainofthoughttoken,cotlength2025}, we note that reasoning ability is strongly influenced by the number of reasoning steps, which are typically considered at the sentence level. In this work, we explored the lexical level, hypothesizing that the number of reasoning actions also affects performance, with an optimal count potentially existing. Using the NLTK toolkit, we extracted and counted verbs from CoT and task-agnostic CoT prompt generations, computing the average number of reasoning verbs per sample across the test dataset. We then analyzed the relationship between prediction accuracy and verb count.

Fig. \ref{fig:action} shows that even task-agnostic CoT prompts can induce question-specific CoT reasoning. Some task-agnostic CoT results approached CoT performance and significantly outperformed zero-shot CoT, indicating that reasoning structure, rather than content, is crucial. We also observed a positive correlation between performance and the average number of reasoning verbs, with an optimal count evident. Below this optimal point, performance improved as the verb count increased; beyond it, performance declined. Additionally, the model exhibited a baseline CoT reasoning ability from pretraining. The weaker performance observed with zero-shot CoT compared to manual CoT may stem from overly divergent reasoning, where excessive actions introduce noise. In contrast, manual CoT maintained balanced reasoning depth, allowing the model to solve problems more effectively with fewer verbs. This observation suggests that the model captures deeper reasoning structures from exemplars.

\begin{tcolorbox}[title=Takeaway]
LLMs leverage ICL to capture fine-grained lexical structures and deeper reasoning patterns from exemplars, while pretrained priors continue to shape the reasoning process.
\end{tcolorbox}

\subsection{RQ2: Can LLMs Override Pretrained Priors Through ICL?}
\label{rq2}
Prior studies~\cite{counterfact,invalid,structurematters} have shown that incorrect reasoning has little effect on LLMs performance. In this section, we examine whether the model’s confidence and accuracy remain stable as the number of exemplars increases. We treat the token generation probability at each time step as a measure of confidence and analyze how this confidence shifts with incorrect reasoning, as well as how accuracy evolves with an increasing number of incorrect exemplars.

% \vspace{-6pt}
\paragraph{\textbf{Method.}}
We adapted methods from~\cite{counterfact,min-etal-2022-rethinking,invalid} to create two contrastive prompt types. The first type, \emph{false-answer CoT prompts}, retains the correct rationale but replaces the final answer. In the Coin Flip dataset, we inverted the answers by replacing ``yes'' with ``no'' and vice versa. In GSM8K, correct answers were replaced with random numbers, while in the Date dataset, correct dates were swapped with random alternatives. The second type, \emph{false-rationale CoT prompts}, modifies the reasoning steps while keeping the final answer correct. In the Coin Flip dataset, key verbs such as ``Flipped'' and ``is'' were replaced with ``not Flipped'' and ``not is''. For the Date dataset, dates were shifted by adding 30 days. In GSM8K, operators were swapped: ``+'' with ``-'', ``-'' with ``+'', ``*'' with ``/'', and ``/'' with ``*''. Due to the limited exemplars provided in~\cite{cot}, we used the small provided set and distilled 40 additional exemplars for each task from the training data using GPT-4~\cite{openai2024gpt4technicalreport}. We then applied the above modifications to create the false-answer and false-rationale CoT prompts.

% \vspace{-20pt}
\begin{figure}[h]
\centering
    \includegraphics[width=0.32\linewidth]{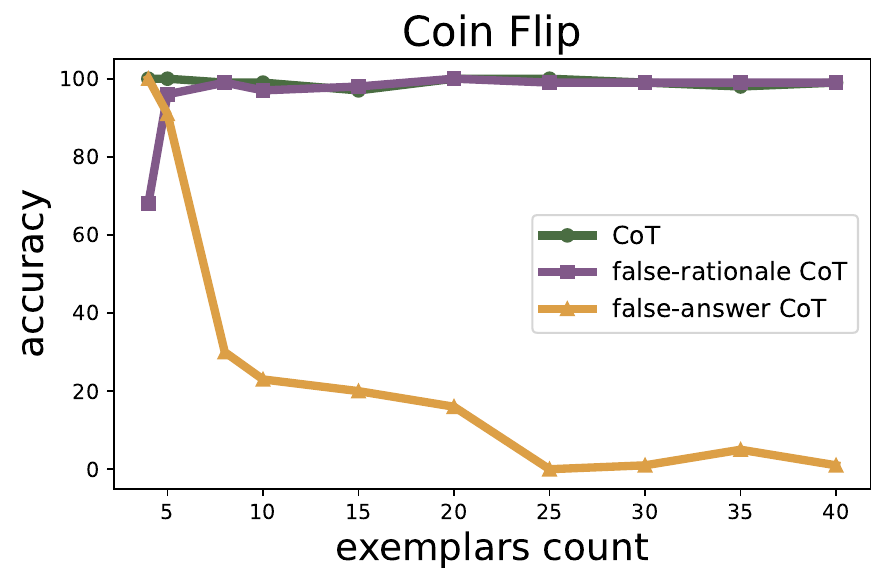}
    \includegraphics[width=0.32\linewidth]{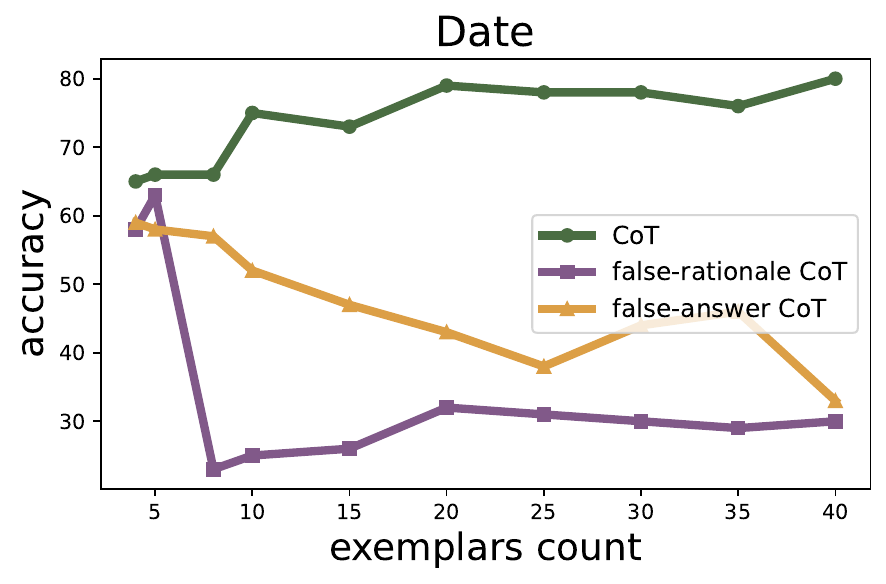}
    \includegraphics[width=0.32\linewidth]{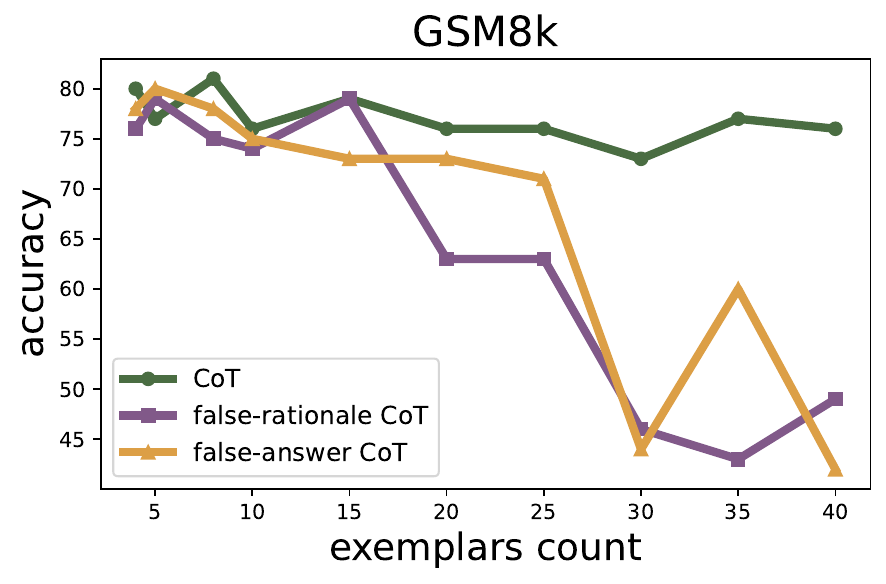}
    \caption{Evolution of test accuracy as the number of \textcolor{darkred}{\textbf{noisy}} (false-answer, false-rationale) exemplars increases.}
    \label{fig:trend}
\end{figure}

% \vspace{-28pt}
\paragraph{\textbf{Analysis.}} 
As shown in Fig.~\ref{fig:trend}, when provided with a small number of exemplars (e.g., 4-shot or 5-shot), the accuracy differences among CoT, false-rationale CoT, and false-answer CoT are minimal. This observation aligns with prior findings that noise in exemplars has little impact on model accuracy during reasoning~\cite{counterfact,invalid}. However, as the number of exemplars increases, these differences become more pronounced. 

We first examine the false-answer CoT prompts. In the Coin Flip task, accuracy declines most sharply as the number of false-answer exemplars increases. This is attributed to its closed-domain nature with a binary label space, where the model tends to learn an input-label mapping that leads to systematic label flipping. In contrast, for the Date and GSM8K tasks, which feature open-domain answer spaces, such mapping is less feasible, resulting in a more gradual decline in accuracy. Nevertheless, at 40-shot, accuracy drops by nearly half, highlighting the cumulative impact of noisy exemplars. Next, we analyze the false-rationale CoT prompts. In the Coin Flip task, accuracy remains relatively stable despite an increasing number of false-rationale exemplars, likely because the small label space allows the model to associate correct answers with certain patterns, minimizing accuracy fluctuations. However, in the Date and GSM8K tasks, noisy rationales significantly impair model reasoning as the exemplar count rises; at 40-shot, accuracy is halved, indicating severe degradation in reasoning ability under large-scale noise. 

Our findings challenge previousous claims. For instance, \cite{wei2023larger} suggested that smaller models cannot learn to flip labels; yet our results show that 8B model exhibit label flipping in closed-domain tasks when using CoT reasoning. Similarly, \cite{invalid} argued that noisy rationales have little effect on reasoning accuracy, we demonstrate that accuracy deteriorates significantly with an increasing number of noisy exemplars. Overall, our analysis reveals an interaction between pretrained priors and ICL in CoT reasoning. In early-shot settings, pretrained priors dominate, stabilizing performance despite noise. As the number of exemplars increases, ICL signals strengthen and shift model decisions based on the provided examples. In closed-domain tasks, this shift leads to systematic label flipping, whereas in open-domain tasks, accuracy declines more gradually yet substantially. These findings underscore the need to balance exemplar quality and quantity to mitigate ICL-induced biases.
% \vspace{-15pt}
\begin{figure}[H]
\centering
    \includegraphics[width=0.9\linewidth, height=0.35\linewidth]{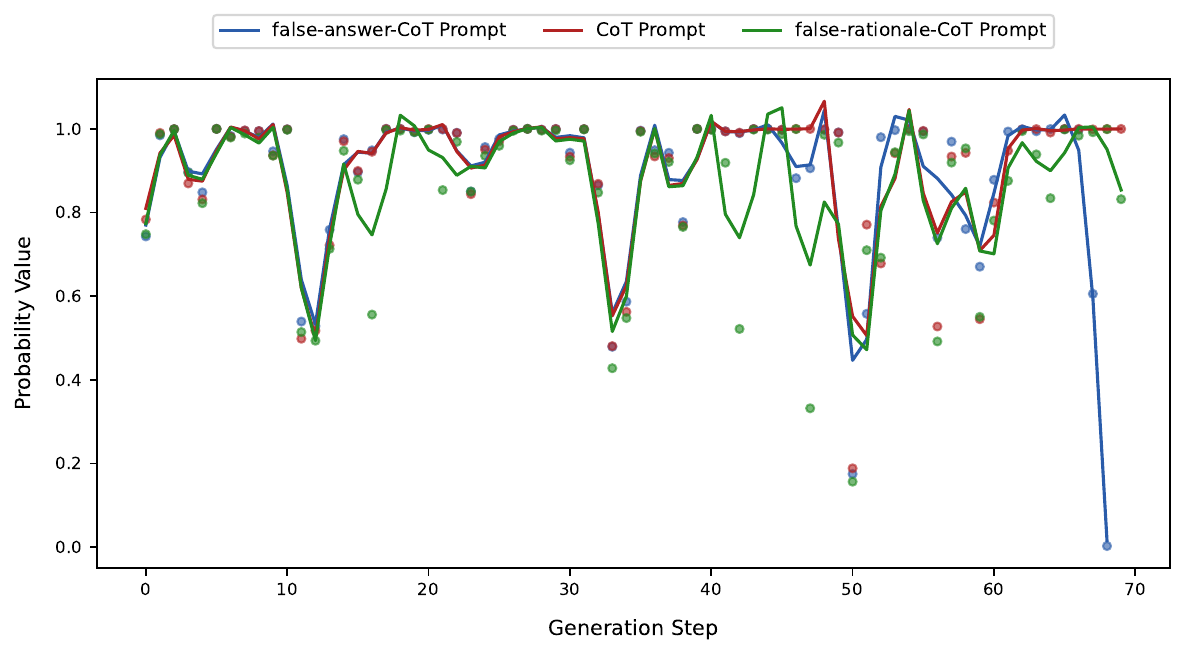}
    \vspace{-10pt}
    \caption{\textcolor{darkyellow}{\textbf{Probabilities evolution}} of model-generated outputs under greedy decoding for three prompt types: \textbf{CoT} prompt, \textbf{false-answer CoT} prompt, and \textbf{false-rationale CoT} prompt.}
    \label{fig:norm}
\end{figure}

% \vspace{-30pt}
\subsubsection{\textit{Case Study.}}
We further investigated the model’s internal mechanisms to understand how false-answer and false-rationale CoT prompts affect reasoning. Under greedy decoding, we recorded the token generation probabilities at each time step to reflect the model’s confidence. Figure~\ref{fig:norm} presents these probabilities over time step for the three prompt types: CoT, false-answer CoT, and false-rationale CoT. The raw probabilities (displayed as scatter points) were obtained by normalizing logits at each generation step, while smoothed curves—derived via Savitzky–Golay filtering\footnote{\url{https://docs.scipy.org/doc/scipy/reference/generated/scipy.signal.savgol_filter.html}}—highlight the overall trends. CoT prompts maintained stable probability values, indicating high confidence. In contrast, false-answer and false-rationale CoT prompts exhibited greater fluctuations, reflecting reduced confidence. This variability suggests that the model detected incorrect reasoning or misleading information, leading to instability. These findings emphasize the role of correct reasoning in reinforcing confidence: CoT prompts provide logical steps that bolster predictions, whereas misleading prompts introduce conflict and uncertainty. From an ICL perspective, correct reasoning enables the model to leverage contextual signals effectively, while misleading prompts undermine stability by contradicting pretrained knowledge. This underscores the importance of well-designed prompts for reliable reasoning.
\vspace{3pt}
\begin{tcolorbox}[title=Takeaway]
Increasing the number of exemplars shifts decision-making from pretrained priors to ICL signals. However, misleading prompts introduce instability, underscoring the importance of high-quality exemplars.
\end{tcolorbox}
\vspace{-8pt}
\subsection{RQ3: Can prompt engineering leverage ICL ability and pretrained knowledge to elicit slow thinking?}
\vspace{-12pt}
\begin{figure}[H]
\centering
    \includegraphics[width=0.6\linewidth, height=0.35\linewidth]{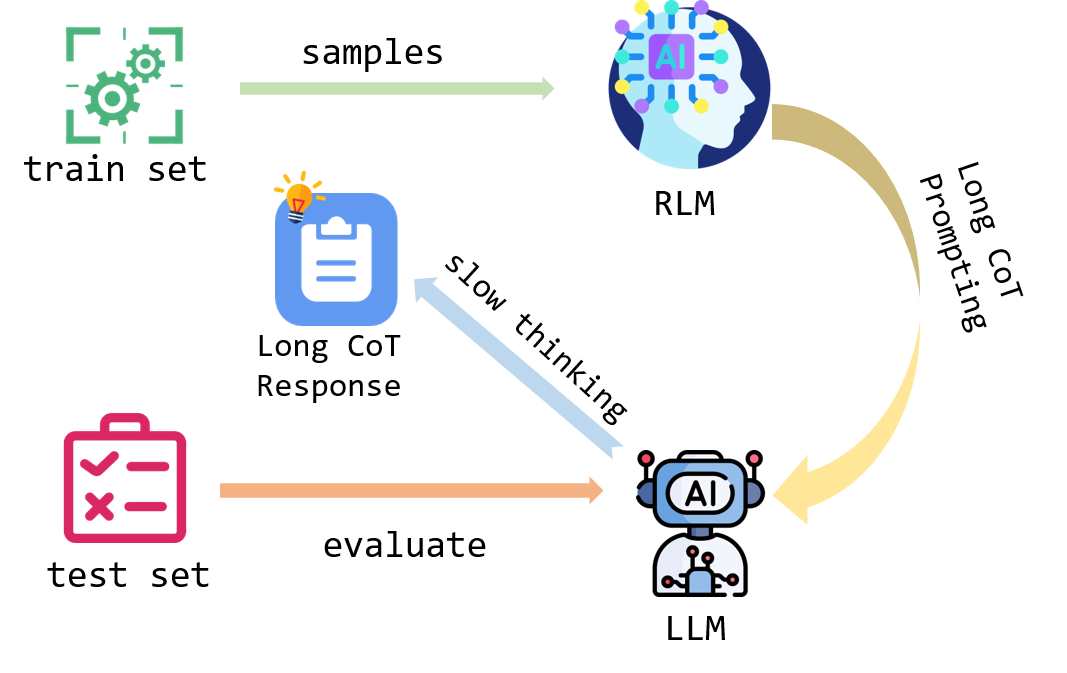}
    \caption{Framework for employing \textcolor{darkyellow}{\textbf{prompt engineering}} to encourage \textcolor{darkgreen}{\textbf{slow thinking}} in LLMs.}
    \label{fig:framework}
\end{figure}
\vspace{-15pt}
Researchers enhance LLMs reasoning abilities by extending the length of CoT, leading to the development of Reasoning Language Models (RLMs). RLMs generate a large number of tokens during inference, a process known as \textit{slow thinking} or \textit{test-time scaling}. The results of two previous experiments indicate that LLMs effectively leverage ICL to adopt the reasoning styles of the exemplars while also utilizing pretrained knowledge. In this section, we investigate whether LLMs can rely on ICL and pretrained knowledge to engage in slow thinking, thereby improving their reasoning performance. As illustrated in Fig.~\ref{fig:framework}, we distilled long CoT prompts from three open-source RLMs: DeepSeek-R1-Distill-Llama-8B, DeepSeek-R1-Distill-Qwen-32B, and QwQ-32B. We then used the long CoT distilled as prompts to guide LLMs (LLaMA3.1-8B, Qwen2.5-32B) in reasoning, examining whether LLMs can learn the slow thinking reasoning approach through prompting. Specifically, we conducted tests on four datasets: GSM8K, MATH-500, Date, and Last Letter Concatenation. The experimental results are shown in Table~\ref{tab:results}.
\begin{table}[h]
    \centering
    \renewcommand{\arraystretch}{1.2}
    \setlength{\tabcolsep}{3.5pt}
    \caption{\textcolor{darkgreen}{Accuracy} and \textcolor{darkyellow}{\textbf{Avg. tokens}} total of different models with various CoT prompts on datasets}
    \resizebox{\textwidth}{!}{
    \begin{tabular}{llcccc}
        \toprule
        \multirow{2}{*}{\textbf{Model}} & \multirow{2}{*}{\textbf{Prompt}} & \multicolumn{4}{c}{\textbf{Accuracy (Avg. tokens total)}} \\
        \cmidrule(lr){3-6}
        & & \textbf{GSM8K} & \textbf{MATH-500} & \textbf{DATE} & \textbf{Last-Letter}\\
        \midrule
        \multirow{4}{*}{LLaMA3.1-8B} 
        & Manual-Short-CoT & 0.2676(92) & 0.112(101) & 0.6016(49) & 0.5504(69)\\
        & DS-LLaMA8B-Long-CoT & \cellcolor{green1}0.4571(364)$\uparrow$ & \cellcolor{green1}0.120(534)$\uparrow$ & \cellcolor{green1}0.6203(252)$\uparrow$ & \cellcolor{green1}0.7217(439)$\uparrow$ \\
        & DS-Qwen32B-Long-CoT & \cellcolor{green2}0.4617(343)$\uparrow$ & \cellcolor{green1}0.125(575)$\uparrow$ & \cellcolor{green1}0.6149(277)$\uparrow$ & \cellcolor{green2}0.7020(484)$\uparrow$ \\
        & QwQ-32B-Long-CoT & \cellcolor{green1}0.3449(650)$\uparrow$ & \cellcolor{green1}0.118(810)$\uparrow$ & \cellcolor{green1}0.6338(252)$\uparrow$ & \cellcolor{green3}0.7560(997)$\uparrow$ \\
        \midrule
        \multirow{4}{*}{LLaMA3.1-8B-Instruction} 
        & Manual-Short-CoT & 0.7695(114) & 0.264(274) & 0.6504(52) & 0.4173(69) \\
        & DS-LLaMA8B-Long-CoT & \cellcolor{green1}0.8248(377)$\uparrow$ & \cellcolor{green1}0.308(765)$\uparrow$ & \cellcolor{green1}0.6991(280)$\uparrow$ & \cellcolor{green3}0.7600(438)$\uparrow$ \\
        & DS-Qwen32B-Long-CoT & \cellcolor{green1}0.7901(383)$\uparrow$ & \cellcolor{green1}0.304(826)$\uparrow$ & \cellcolor{red1}0.6449(322)$\downarrow$ & \cellcolor{green3}0.8830(494)$\uparrow$ \\
        & QwQ-32B-Long-CoT & \cellcolor{red1}0.6202(702)$\downarrow$ & \cellcolor{green1}0.298(767)$\uparrow$ & \cellcolor{green1}0.6856(368)$\uparrow$ & \cellcolor{green3}0.8487(589)$\uparrow$ \\
        \midrule
        \multirow{4}{*}{Qwen2.5-32B} 
        & Manual-Short-CoT & 0.8104(150) & 0.310(407) & 0.6991(62) & 0.7903(72) \\
        & DS-LLaMA8B-Long-CoT & \cellcolor{green1}0.8195(425)$\uparrow$ & \cellcolor{green3}0.410(619)$\uparrow$ & \cellcolor{green3}0.7775(315)$\uparrow$ & \cellcolor{green1}0.8185(436)$\uparrow$ \\
        & DS-Qwen32B-Long-CoT & \cellcolor{green2}0.8599(440)$\uparrow$ & \cellcolor{green2}0.392(698)$\uparrow$ & \cellcolor{green2}0.7667(368)$\uparrow$ & \cellcolor{green1}0.8302(472)$\uparrow$ \\
        & QwQ-32B-Long-CoT & \cellcolor{green3}0.8777(511)$\uparrow$ & \cellcolor{green3}0.425(1082)$\uparrow$ & \cellcolor{green2}0.7639(427)$\uparrow$ & \cellcolor{green1}0.8161(631)$\uparrow$ \\
        \midrule
        \multirow{4}{*}{Qwen2.5-32B-Instruction} 
        & Manual-Short-CoT & 0.8316(117) & 0.590(415) & 0.8292(108) & 0.7802(192) \\
        & DS-LLaMA8B-Long-CoT & \cellcolor{red1}0.8221(401)$\downarrow$ & \cellcolor{green1}0.602(688)$\uparrow$ & \cellcolor{green3}0.8886(379)$\uparrow$ & \cellcolor{green3}0.9052(1032)$\uparrow$ \\
        & DS-Qwen32B-Long-CoT & \cellcolor{green3}0.8945(681)$\uparrow$ & \cellcolor{green2}0.626(586)$\uparrow$ & \cellcolor{green2}0.8721(344)$\uparrow$ & \cellcolor{green2}0.8649(1058)$\uparrow$ \\
        & QwQ-32B-Long-CoT & \cellcolor{green1}0.8529(893)$\uparrow$ & \cellcolor{green3}0.638(1305)$\uparrow$ & \cellcolor{green2}0.8723(390)$\uparrow$ & \cellcolor{green1}0.8407(1105)$\uparrow$\\
        \bottomrule
    \end{tabular}
    }
    \label{tab:results}
\end{table}
% \vspace{-20pt}
By checking the generated content, we find that the model effectively emulates the reflective, retrospective, and summarization-based slow thinking of long CoT, leading to improved performance on downstream tasks. This effect is particularly pronounced in instruct models due to their strong instruction following capabilities.
The data in the Table~\ref{tab:results} indicates an optimal CoT reasoning length for the model, influenced by model capacity and task difficulty. This result is consistent with the findings of~\cite{cotlength2025,yang2025thinkingoptimalscalingtesttimecompute}. For instance, on GSM8K with the LLaMA3.1-8B-Instruct, performance declines when CoT length exceeds a certain point. In the 8B model, shorter CoT length yields strong performance, but longer length leads to degradation. Similarly, in the Qwen2.5-32B model, the optimal CoT length increases with model size, and shorter CoT lengths result in poorer reasoning performance.

\begin{tcolorbox}[title=Takeaway]
Leveraging ICL and pretrained knowledge, LLMs can adopt slow thinking through prompt engineering, effectively generating long CoT reasoning.
\end{tcolorbox}
% \vspace{-15pt}
\section{Related Works}
\vspace{-2pt}
\paragraph{In-Context Learning.}Recent studies have examined the role of labels in ICL.~\cite{yoo-etal-2022-ground} revisit label randomization and report significant variance across tasks and models. ~\cite{pan-etal-2023-context} distinguish between label-independent and label-dependent learning by substituting labels with arbitrary tokens.~\cite{wei2023larger} demonstrate that smaller models struggle with ICL when labels are replaced. However, these studies overlook probabilistic metrics, potentially underestimating the impact of label modifications. For instance, ~\cite{pan-etal-2023-context} suggest that the performance gap between random and default labels is insignificant for small models, while~\cite{wei2023larger} argue that large models can override pretraining priors in context, whereas small models fail to adjust to flipped labels—claims that our findings in Section~\ref{rq2} challenge.
\vspace{-10pt}
\paragraph{Chain-of-Thought.}CoT developments have significantly improved performance, particularly in complex reasoning tasks such as arithmetic and commonsense reasoning~\cite{cot}. This success has led to research on  Least-to-Most prompting~\cite{zhou2023leasttomost}, and specialized techniques such as Program-of-Thought~\cite{chen2023program}, Contrastive Chain-of-Thought~\cite{contrastive}. Research has examined various factors influencing CoT to enhance prompt design. The use of diverse examples has been found to complement performance~\cite{ye-etal-2023-complementary}. Maintaining logical coherence, even in the presence of errors, retains performance~\cite{contrastive,wang-etal-2023-towards,invalid,structurematters}. However, understanding why CoT is effective remains limited. Some studies suggest that reasoning abilities arise during pretraining, with exemplars guiding generation~\cite{wang2024chainofthought,contrastive,saparov2023language}, though explicit empirical support is lacking. Other research argues that intermediate steps serve as templates rather than aiding task-solving [32]. Further findings indicate that imitation improves style adherence but not factuality or problem-solving~\cite{madaan2022textpatternseffectivechain}. 

Unlike the above studies, our work explores CoT mechanisms from the perspectives of ICL and pretrained priors, highlighting their interplay for a cohesive understanding.
% \vspace{-16pt}
\section{Conclusion}
% \vspace{-6pt}
Our study demonstrates that LLMs, through ICL, not only capture lexicon-level reasoning structures from CoT exemplars but also internalize deeper reasoning logic. Providing sufficient exemplars shifts the model's decision-making from pre-trained priors to ICL signals. Conversely, misleading prompts induce instability, highlighting the crucial role of exemplar quality. Moreover, by leveraging ICL and pretrained knowledge, prompt engineering enables models to partially emulate slow thinking, offering a promising path for self-evolution.
\vspace{-6pt}
\section{Limitation}
\vspace{-6pt}
This study explores the mechanism of CoT reasoning from the perspective of the interplay between ICL and pretrained priors, further examining the potential of inducing slow thinking through prompt engineering. However, our experiments are limited to a few datasets, mainly in mathematics, leaving more complex reasoning tasks to be tested. Additionally, the impact of different levels of noise on CoT reasoning requires further investigation. Finally, our use of greedy decoding for long CoT often resulted in endless repetitions, indicating the need to explore alternative model settings for better performance.
\bibliographystyle{splncs04}
\bibliography{reference}
\end{document}